\documentclass[10pt,twocolumn,letterpaper]{article}

\usepackage[pagenumbers]{iccv} 

\definecolor{iccvblue}{rgb}{0.21,0.49,0.74}
\usepackage[pagebackref,breaklinks,colorlinks,allcolors=iccvblue]{hyperref}

\usepackage{graphicx}	
\usepackage{amsmath}	
\usepackage{amssymb}	
\usepackage{booktabs}
\usepackage{times}
\usepackage{microtype}
\usepackage{epsfig}
\usepackage{caption}
\usepackage{float}
\usepackage{placeins}
\usepackage{color, colortbl}
\usepackage{stfloats}
\usepackage{enumitem}
\usepackage{tabularx}
\usepackage{booktabs}
\usepackage{xstring}
\usepackage{multirow}
\usepackage{xspace}
\usepackage{url}
\usepackage{subcaption}
\usepackage[hang,flushmargin]{footmisc}
\usepackage{makecell}
\usepackage{multicol}
\usepackage{balance}

\usepackage{algorithm}
\usepackage{algpseudocode}

\usepackage{adjustbox}

\algrenewcommand\algorithmicrequire{\textbf{Input:}}
\algrenewcommand\algorithmicensure{\textbf{Output:}}

\def\paperID{1335} 
\def\confName{ICCV}
\def\confYear{2025}

\title{HiLLIE: Human-in-the-Loop Training for Low-Light Image Enhancement}

\author{\hspace{-0.5cm}
Xiaorui Zhao$^1$\quad Xinyue Zhou$^1$\quad
Peibei Cao$^2$\quad Junyu Lou$^1$\quad  Shuhang Gu$^1$\thanks{Corresponding author.}\\
\hspace{-0.5cm}
$^1$University of Electronic Science and Technology of China \hspace{0pt}\\
$^2$ Nanjing University of Information Science and Technology\\
\hspace{-0.5cm}
{\tt\small \{zzzhaoxiaorui, shuhanggu\}@gmail.com}}

\begin{document}
\maketitle
\begin{abstract}
Developing effective approaches to generate enhanced results that align well with human visual preferences for high-quality well-lit images remains a challenge in low-light image enhancement (LLIE).
In this paper, we propose a human-in-the-loop LLIE training framework that improves the visual quality of unsupervised LLIE model outputs through iterative training stages, named HiLLIE.
At each stage, we introduce human guidance into the training process through efficient visual quality annotations of enhanced outputs.
Subsequently, we employ a tailored Image Quality Assessment (IQA) model to learn human visual preferences encoded in the acquired labels, which is then utilized to guide the training process of an enhancement model.
With only a small amount of pairwise ranking annotations required at each stage, our approach continually improves the IQA model's capability to simulate human visual assessment of enhanced outputs, thus leading to visually appealing LLIE results.
Extensive experiments demonstrate that our approach significantly improves unsupervised LLIE model performance in terms of both quantitative and qualitative performance.
The code and collected ranking dataset will be available at \href{https://github.com/LabShuHangGU/HiLLIE}{https://github.com/LabShuHangGU/HiLLIE}.
\end{abstract}    
\section{Introduction}
\label{sec:intro}
The task of low-light image enhancement (LLIE) aims to improve the visibility of underexposed images taken in low or uneven lighting conditions. 
Since inadequate illumination severely degrades the visual structures of images in digital photography, LLIE is of practical importance and has been a thriving area of research over the past few years.

Traditional LLIE methods~\cite{tradition_HE1,tradition_HE2,tradition_HE3,tradition_re1,tradition_re2,tradition_re3} mainly adopt hand-crafted priors to improve image contrast and illumination. However, these priors only possess limited adaptability, leading to unrealistic results such as over- and under-exposure, color cast, and detail loss.
Benefiting from the rapid development of deep-learning, supervised LLIE methods~\cite{super_re2_KID,super_re3_KID++,super_re4_uretinexnet,super_retinexformer,super_semantic1,super_quantization,super_snr,super_lap_DSLR} can produce well-lit enhanced results with great preservation of image details. 
The success of these methods heavily relies on large-scale paired training data, which naturally infuses human visual preferences into model training through human retouched ground-truth images.
For instance, MIT-Adobe FiveK~\cite{MIT5K} is a large photo enhancement dataset that provides finely retouched ground-truth images processed by five human experts.
By utilizing images adjusted by professional experts as ground-truths for supervised training, enhancement models can generate results consistent with human visual preferences.
However, due to the inconsistency of human evaluation criteria, even the same expert may produce different retouched results when manually adjusting the same image multiple times, leading to unstable learning targets and difficulties in model training.
Also, it is labor-intensive and time-consuming to collect and process a large amount of paired training data in real-world scenarios.

\begin{figure}[t]
    \centering
    \includegraphics[width=0.99\linewidth]{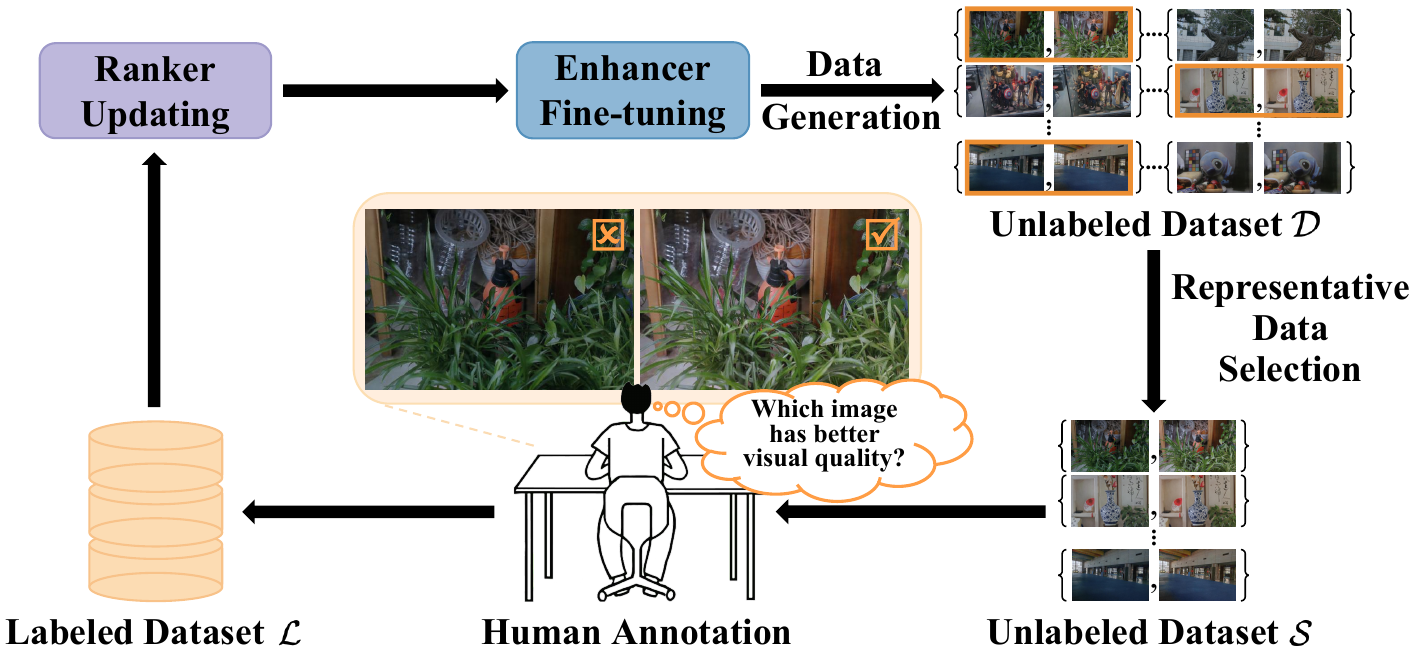}
    \caption{
    Illustration of the HiLLIE training framework.
    }
    \label{fig:HiLLIE_framework}
\end{figure}

To alleviate this data limitation, unsupervised approaches leverage unpaired low- and normal-light training data to enhance the visual appearance of underexposed images through distribution alignment between enhanced outputs and real-world well-exposed  images~\cite{unsuper_clip_lit,unsuper_enlightengan,unsuper_lightendiffusion,unsuper_NeRco}.
Meanwhile, zero-reference methods train with either low- or normal-light images, exploiting physical priors~\cite{zeroref_quadprior} and ideal assumptions~\cite{zeroref_zero-DCE,zeroref_zero-DCE++} to improve image visibility in low-light scenes.
Although these methods can achieve notable improvements in overall image illumination and perceptual quality, neither distribution alignment nor image prior exploration can generate enhanced results that match human visual preferences, since they lack direct human guidance during model optimization.
A relevant approach to ours is~\cite{band}, which aims to generate visually appealing enhanced results by applying a pretrained general Image Quality Assessment (IQA) model as quality guidance.
However, general IQA approaches often fall short in accurately assessing the visual quality of LLIE model outputs.
This limitation stems from the fact that most IQA models are designed for universal scenarios, whose training data cover a diverse range of image degradations instead of focusing on underexposed image impairments.

In this work, we propose a human-in-the-loop LLIE training framework, HiLLIE, which directly integrates human perceptual preferences into the training process of an unsupervised LLIE model with the help of a tailored IQA model.
As shown in Fig.~\ref{fig:HiLLIE_framework}, our framework iteratively trains the LLIE model (termed enhancer) and the IQA model (termed ranker) in an alternating manner over multiple stages.
In each stage, by applying the most recently updated enhancer to its training low-light inputs, we generate enhanced images which form content-identical pairs with those generated from the previous stage.
Then, we employ a dense data selection strategy to identify the most representative image pairs for subsequent human annotation.
By requiring human subjects to label only a small subset of image pairs, we achieve efficient human annotation with a small amount of labeling effort.
Moreover, asking human subjects to perform pairwise quality comparisons between two enhanced results of the same content within each image pair, instead of assigning a quality score to single image, significantly reduces the potential labeling bias caused by annotation instability.
After that, all accumulated data and labels are utilized to train the ranker model, enabling it to accurately simulate human visual assessment of enhancer outputs. 
Finally, we adopt a ranker loss computed by the ranker model to guide the fine-tuning process of enhancer model.
With only a small amount of annotation cost required at each iterative stage, ranker can progressively improve its capability to capture human visual preferences, thereby guiding enhancer to generate high-quality LLIE outputs with human-desired characteristics through the ranker loss.

To the best of our knowledge, this is the first attempt to introduce the idea of human-in-the-loop into the training process of LLIE models. 
Extensive experiments show that our method significantly improves unsupervised LLIE model performance and generates enhanced results with superior visual quality.
Our contributions are as follows:
\begin{itemize}
  \item We propose a human-in-the loop training framework for unsupervised LLIE models, which successfully incorporates human visual preferences into the model training process with only small-scale annotation effort, enabling the generation of human-desired well-lit outputs.

  \item We provide a specific implementation of our framework with a novel rank-based data annotation approach, where a small fixed number of content-identical image pairs are annotated by human subjects at each iterative stage. The collected data and labels not only facilitate the continuous improvement of ranker model but also serve as a valuable dataset for training IQA models specifically tailored for evaluating LLIE model performance.

  \item Extensive experimental results demonstrate that our approach outperforms current leading unsupervised LLIE methods in both subjective visual assessments and objective metric evaluations.

\end{itemize}
\vspace{-1mm}
\section{Related Work}
\label{sec:related}

\subsection{Low-Light Image Enhancement}
Recent years have witnessed a surge in research efforts dedicated to the enhancement of low-light images, with the goal of producing well-exposed, visually appealing enhanced results.
Conventional LLIE approaches primarily rely on manually designed optimization algorithms. These methods fall into two categories: histogram equalization-based approaches~\cite{tradition_HE1,tradition_HE2,tradition_HE3}
that focus on adjusting image contrast through histogram manipulation, and retinex-based methods~\cite{tradition_re1,tradition_re2,tradition_re3,tradition_re4,tradition_re5} that operate by decomposing and modifying image illumination characteristics.

LLNet~\cite{super_llnet} pioneered the introduction of deep learning into the LLIE field, proposing a deep autoencoder-based model to generate normal-light enhanced results.
Since then, there has been an explosion in the number of DNN-based supervised LLIE methods that leverage advanced network architectures~\cite{super_Flow,super_retinexformer,super_diff} and draw inspiration from established theories~\cite{super_re1,super_re2_KID,super_re3_KID++,super_re4_uretinexnet, super_semantic1,super_fft1}, aiming to better learn the mapping from low-light to normal-light images.
To mitigate the dependence of supervised approaches on extensive paired training data, several unsupervised methods have been proposed. 
EnlightenGAN~\cite{unsuper_enlightengan} applies adversarial training to adjust the local and global features of low-light images. 
PairLIE~\cite{unsuper_PairLIE} introduces a novel data paradigm that exploits priors from paired low-light images of the same scene. 
With the help of rich visual-language prior of Contrastive Language-Image Pre-Training (CLIP) model~\cite{CLIP}, CLIP-LIT~\cite{unsuper_clip_lit} leverages prompt learning and contrastive learning to establish adaptive image quality standards for model fitting. 
Recently, LightenDiffusion~\cite{unsuper_lightendiffusion} integrates retinex theory with the generative power of diffusion model for unsupervised enhancement. 
By aligning the distributions between model outputs and natural normal-light images, unsupervised LLIE models can generate enhanced results with promising visual quality.
Several zero-reference LLIE approaches~\cite{zeroref_RUAS,zeroref_sci,zeroref_zero-DCE,zeroref_zero-DCE++,zeroref_quadprior} have also been developed to further eliminate the reliance on image pairs by only utilizing low- or normal-light images for training, achieving better generalization in real-world scenes.
Notably, most existing unsupervised and zero-reference LLIE methods have not managed to effectively incorporate human perceptual preferences into their optimization process, indicating potential for enhancing results towards better human visual perception.

\subsection{Human-in-the-Loop Training in Computer Vision}
Human-in-the-loop~\cite{human_book} is a training strategy based on interactions between human and machine learning algorithms. This approach enhances model accuracy and generalization ability by incorporating human feedback throughout the training cycle.
During the training phase, human-in-the-loop strategy employs active learning methods~\cite{active1,active2} to select the most valuable data samples for human annotation, thereby improving labeling efficiency.
Transfer learning~\cite{transfer_learning} is also used in human-in-the-loop training frameworks, which mitigates the cold-start problem by leveraging pre-trained models for new tasks instead of training from scratch.
In practice, human-in-the-loop training strategies have been widely adopted across various high-level computer vision tasks, including instance segmentation~\cite{human_instance_seg}, semantic segmentation~\cite{human_semantic_seg1,human_semantic_seg2,human_semantic_seg3}, person re-identification~\cite{human_person_re_identi}, anomaly detection~\cite{human_anomaly}, and text-to-image generation~\cite{human_T2I}.
To the best of our knowledge, we are the first to adopt the human-in-the-loop training approach for LLIE models.
\section{Methodology}
\label{sec:method}

\subsection{Overview of HiLLIE}
In this section, we present our proposed HiLLIE training framework, a novel training strategy that optimizes two key components: a low-light image enhancer model and an image quality ranker model.
This integrated process progressively improves ranker's capability to simulate human visual assessment of enhancer outputs, which in turn guides the enhancer through ranker loss to generate enhanced results that better align with human visual preferences.

In our implementation, we validate our HiLLIE training strategy on LightenDiffusion~\cite{unsuper_lightendiffusion}, a state-of-the-art unsupervised LLIE approach trained using two sets of unpaired low-light images $\mathcal{X}\!=\!\{ x_{i}\}_{i = 1}^{|\mathcal{X}|}$
and normal-light images $\mathcal{Y} = \{ y_{i}\}_{i = 1}^{|\mathcal{Y}|}$.
The proposed $N$-stage HiLLIE training framework involves an initialization phase and subsequent iterative ranker-enhancer training cycles:

\noindent \textbf{Phase 1: Model Initialization ($n=0$).}  We begin our HiLLIE framework by pretraining an unsupervised LLIE model as the initial enhancer model $f^{(0)}$, ensuring relatively good enhanced results in subsequent fine-tuning cycles. 
Moreover, to address the challenge of insufficient data and labels in the early stages, we train an NIQE-oriented ranker model $g^{(0)}$, which serves as a robust initialization for human-oriented ranker models. 
This $g^{(0)}$ is trained using enhanced images generated by intermediate checkpoints saved during the enhancer pretraining process and the final model $f^{(0)}$ on low-light training inputs, along with their corresponding NIQE scores.
After that, we fine-tune $f^{(0)}$ to obtain $f^{(1)}$ by leveraging $g^{(0)}$ through a ranker loss, along with a fidelity loss to preserve content consistency.

\setlength{\textfloatsep}{2em} 
\begin{algorithm}[t]
\caption{Human-in-the-Loop Low-Light Image Enhancement (HiLLIE) Training Framework}
\label{alg:hillie}
\begin{algorithmic}[1]
\Require unpaired low-light images $\mathcal{X}\!=\!\{ x_{i}\}_{i = 1}^{|\mathcal{X}|}$ and normal-light images $\mathcal{Y} = \{ y_{i}\}_{i = 1}^{|\mathcal{Y}|}$
\Ensure Improved enhancer model $f^{(N)}$
\For{$n = 0$ to $N-1$}
    \State $\mathcal{D}\leftarrow\emptyset$, $\mathcal{S}\leftarrow\emptyset$
    \If{$n = 0$}
        \State Initialize enhancer to obtain $f^{(0)}$
        \State Train NIQE-oriented ranker $g^{(0)}$
        \State Fine-tune $f^{(0)}$ with $g^{(0)}$ to obtain $f^{(1)}$
    \Else
        \State Generate $\hat{\mathcal{Y}}^{(n)}$ using $f^{(n)}$ on $\mathcal{X}$ and form $\mathcal{D}^{(n)}$
        \State $\mathcal{D}\leftarrow\mathcal{D}\cup\mathcal{D}^{(n)}$
        \State Select $\mathcal{S}^{(n)}$ according to Eq.~(\ref{eq:1})
        \State $\mathcal{S}\leftarrow\mathcal{S}\cup\mathcal{S}^{(n)}$
        \State Collect human-annotated labels for $\mathcal{S}^{(n)}$
        \State Train human-oriented ranker $g^{(n)}$ using $\mathcal{S}$
        \State Fine-tune $f^{(n)}$ with $g^{(n)}$ to obtain $f^{(n+1)}$
    \EndIf
\EndFor
\State \Return $f^{(N)}$
\end{algorithmic}
\end{algorithm}

\noindent \textbf{Phase 2: Iterative Training Cycles ($n\geq1$).}  
At each stage $n$, the process begins with applying $f^{(n)}$ to the low-light training inputs, generating enhanced images $\hat{\mathcal{Y}}^{(n)} = \{ \hat{y}_{i}^{(n)}\}_{i = 1}^{|\hat{\mathcal{Y}}^{(n)}|}$. Images derived from the same input in adjacent stages are then matched and stored as pairs $\hat{\mathcal{D}}^{(n)} = \{ (\hat{y}_i^{(n-1)}$, $\hat{y}_i^{(n)})\}_{i = 1}^{|\hat{\mathcal{Y}}^{(n)}|}$ in dataset $\mathcal{D}$.
Following this, we implement a dense data selection strategy, choosing top-$k$ image pairs from $\mathcal{D}^{(n)}$ with the largest discrepancy in predicted values by $g^{(n-1)}$. These selected pairs, denoted as $\mathcal{S}^{(n)}$, form a subset $\mathcal{S}$ of $\mathcal{D}$, enabling efficient human evaluation on a small number of image pairs at each stage. 
Next, human annotators compare the overall image quality between the two images in each pair within $\mathcal{S}^{(n)}$, providing ranking labels that reflect human visual perception. 
To avoid potential model forgetting problem, all accumulated data and its corresponding labels are stored in dataset $\mathcal{L}$ and utilized to update the ranker model. 
During this process, both images in each pair are processed through an identical network architecture to generate corresponding predicted quality scores. These scores are then used to compute the margin-ranking loss, guiding ranker to learn the quality ranking relationships encoded in the labels and obtain $g^{(n)}$. 
Through multiple training stages, the ranker model progressively improves its capability to simulate human visual assessment of enhancer outputs.
This enables $g^{(n)}$ to guide the fine-tuning of $f^{(n)}$ to obtain $f^{(n+1)}$ through ranker loss, which generatively assesses and improves the perceptual quality of LLIE results. We also apply the same content fidelity loss as in the model initialization phase to ensure training stability.

The proposed framework is summarized in Alg.~\ref{alg:hillie}. 
In the following sections, we would delve into the detailed designs of ranker and enhancer models in our HiLLIE framework.

\subsection{Image Quality Ranker}
\subsubsection{Ranking dataset}
\label{sec:ranking dataset}
To develop an IQA model that accurately evaluates the visual quality of enhancer results, the training data and labels are of critical importance. This section outlines our process to collect and label images for training the IQA model, which we refer to as ranker.
As previously mentioned, our ranking dataset is accumulated throughout the human-in-the-loop training process. At each stage, we generate new ranking data by applying the latest enhancer model to low-light training inputs. Human anntators then compare and rank these new outputs against those generated from the previous stage, providing quality labels for the ranking data. 
Notably, compared to general IQA methods whose training data contains various types of image degradation and may have unreliable quality assessment on LLIE model outputs, our ranking dataset exclusively contains enhanced images generated by enhancer models at different training stages, making ranker a tailored IQA model for more accurate assessment of enhancer output quality.

To obtain enhanced images with acceptable quality as ranking data at the initial training stage, we first need to acquire a reasonably pretrained LLIE model, denoted as $f^{(0)}$.
Given the limited amount of ranking data and human-annotated labels initially available, relying solely on this dataset may lead to suboptimal convergence and limited ranker performance.
To address this issue, we choose to start with training an NIQE-oriented ranker model using NIQE labels before incorporating human-annotated labels. 
Specifically, we utilize all intermediate models saved during the enhancer pretraining process, along with the final $f^{(0)}$, to inference on the training low-light inputs. This yields a diverse set of approximately $2,000$ enhanced images for ranker training.
Following this, we employ NIQE~\cite{NIQE}, a no-reference IQA perceptual metric, to assign a discrete quality score to each image, which serves as the corresponding NIQE label.
This NIQE scoring enables us to establish quality rankings between any pair of enhanced versions derived from the same input image, providing ample training data and labels for the NIQE-oriented ranker.
We then train ranker with all the data and corresponding NIQE labels to obtain ranker model $g^{(0)}$.
Notably, since we aim to develop a ranker that emulates human visual assessment of enhancer outputs rather than specific NIQA metrics, these NIQE labels are not utilized in later ranker training stages. Instead, we only apply $g^{(0)}$ as a pretrained model to facilitate a favorable initialization for training the human-oriented ranker models.

After fine-tuning $f^{(0)}$ using $g^{(0)}$ to obtain $f^{(1)}$, we commence the annotation of human ranking labels. 
In stage $n$ ($n\geq1$), we first apply the latest enhancer $f^{(n)}$ to the low-light training inputs and obtain enhanced images $\hat{\mathcal{Y}}^{(n)}$.
Image pairs are then formed by one-to-one content-identical correspondence between $\hat{\mathcal{Y}}^{(n-1)}$ and $\hat{\mathcal{Y}}^{(n)}$ and are stored in a large-scale dataset $\mathcal{D}$.
To ensure that only a small and consistent number of image pairs require annotation at each stage, we employ a dense data selection method. Among all the image pairs in $\mathcal{D}^{(n)}$, we select $k$ image pairs with the largest discrepancy in quality scores predicted by $g^{(n-1)}$ to form a smaller set $\mathcal{S}^{(n)}\subset\mathcal{D}^{(n)}$, as given below:
\begin{equation}
\small
\label{eq:1}
    \mathcal{S}^{(n)}\!\!=\!\Big\{\!(\hat{y}_{i}^{(n\!-\!1)}\!,\!\hat{y}_{i}^{(n)})\!\in\!T_k\Big(\!\big|g^{(n\!-\!1)}(\hat{y}_{i}^{(n\!-\!1)})\!-\!g^{(n\!-\!1)}(\hat{y}_{i}^{(n)})\big|\!\Big)\!\Big\}
\end{equation}
where $(\hat{y}_{i}^{(n\!-\!1)}, \hat{y}_{i}^{(n)})\!\in\!\mathcal{D}^{(n)}$ and $T_k$ denotes the operator that selects the top-$k$ elements.
The selected set $\mathcal{S}^{(n)}$ facilitates efficient labeling by human annotators.
To address model forgetting problem during ranker's iterative refinement, we utilize all collected ranking data and its corresponding labels when training $g^{(n)}$.

During the human annotation process, instead of imposing rigid criteria for human assessment, we instruct human annotators to compare each image pair based on their first, overall visual impression, without fixating on specific details such as noise or compression artifacts.
This strategy helps to capture labels that better reflect holistic human visual perception, which enables us to train a ranker model that closely aligns with human judgment of enhancer outputs.
To increase the reliability of the collected ranking labels, each image pair is assessed by three independent annotators. For a given pair of images with identical content, each annotator assigns a label of either 0 or 1 to one of the images, resulting in two possible label combinations: 0/1 or 1/0. Similar to NIQE metric, lower value indicates better perceived image quality. After all human annotations are completed, the label combination that receives the majority of votes is selected as the final ranking label for the image pair. Notably, the collected data and labels not only support the continual refinement of the ranker model in our HiLLIE framework, but also provide a valuable dataset for training IQA models specifically designed for LLIE task.

\subsubsection{Quality Ranking Network Training}
In each training stage, once the ranking dataset is prepared, the proposed ranker is trained to learn the ranking orders of image pairs within the dataset.
The ranker model adopts a VGG-like architecture~\cite{VGG}, which consists of multiple convolutional blocks incorporating convolutional layers, batch normalization~\cite{batchnorm}, and LeakyReLU activation~\cite{leakyrelu}. 
After the convolutional blocks, a global average pooling layer is applied to summarize spatial information, allowing the network to handle varying input sizes.
The resulting feature vector is then fed into a classifier consisting of two fully connected layers with a LeakyReLU activation in between. The output layer produces a scalar value, which represents the relative quality of the input image.

Since we only conduct rank labeling for each image pair instead of assigning an absolute quality score for every image, we draw upon the idea of learning to rank~\cite{rankiqa,ranksrgan,underwaterranker} to train the ranker model.
For each image pair $\{s_n, s_m\}$ in set $\mathcal{S}$, both images are processed through an identical ranker architecture, yielding corresponding predicted scores $\{p_n, p_m\}$ from the model. 
Following this, the predicted scores are utilized to compute the margin-ranking loss, which guides ranker to generate scores that preserve the relative quality order encoded in the ranking labels.
The margin-ranking loss is defined as:
\begin{equation}
    \ell(p_n, p_m) = 
    \begin{cases}
        \mathrm{max}(0, (p_m-p_n) + \epsilon), & r_n \geq r_m \\
        \mathrm{max}(0, (p_n-p_m) + \epsilon), & r_n < r_m
    \end{cases},
\end{equation}
where $r_n$ and $r_m$ represent the ranking labels of $s_n$ and $s_m$, respectively.
$\epsilon$ is a margin parameter that enforces a minimum distance between the predicted scores of paired images. 
Notably, given that we do not normalize the scalar outputs of the ranker model, these predicted values should not be interpreted as absolute quality scores for the images. Instead, they function solely as relative quality indicators that reflect the comparative relationships of visual quality between different images.

\subsection{Low-Light Image Enhancer}
\label{sec:enhancer loss}
In this section, we provide a detailed explanation of how we utilize the newly updated ranker to fine-tune the enhancer at each training stage of HiLLIE. 
In principle, our proposed HiLLIE training framework can be applied to fine-tune any LLIE model, helping it produce enhanced results that better align with human visual quality preferences. 
In this study, we employ LightenDiffusion~\cite{unsuper_lightendiffusion} as our baseline network, a state-of-the-art unsupervised LLIE approach that achieves competitive results on established low-light benchmarks. 
After pretraining the baseline network according to its original settings, we obtain the final pretrained enhancer model, $f^{(0)}$, which serves as the starting point for our HiLLIE method.

By continuously collecting ranking labels from human subjects for enhanced image pairs and utilizing all ranking data and labels to train the ranker, we progressively improve its ability to simulate human visual assessment of enhancer output quality. 
This allows us to employ the differentiable ranker as a training criterion, enabling accurate quality evaluation of enhancer outputs during training. 
Specifically, in stage $n$ of the HiLLIE training framework, we fine-tune $f^{(n)}$ to obtain $f^{(n+1)}$ using the ranker loss~\cite{ranksrgan, underwaterranker} calculated by $g^{(n)}$, which guides the enhancer to produce perceptually improved results that better match human visual preferences.
The formulation of ranker loss is expressed as follows:
\begin{equation}
\ell_\mathrm{R}=\mathrm{Sigmoid}(g^{(n)}(\hat{y}_{i}^{(n+1)})),
\end{equation}
where $\hat{y}_{i}^{(n+1)}$ is the enhancer output, and $g^{(n)}(\hat{y}_{i}^{(n+1)})$ is its quality score predicted by $g^{(n)}$. 

To reinforce training stability and avoid significant deviations during the fine-tuning process of enhancer model, we incorporate an additional loss to constrain the content of model outputs. 
Given our unpaired training setting, we adopt the self-feature preserving loss~\cite{unsuper_enlightengan}, which is designed based on the observation that VGG model classifications exhibit relative insensitivity to manipulations of image pixel intensity ranges~\cite{sfp2,unsuper_enlightengan}.
In our approach, we impose a content fidelity loss between the extracted feature maps $\phi$ of enhancer outputs at adjacent stages, which is defined as:
\begin{equation}
\ell_{\mathrm{con}}=\sum_i||\phi(\hat{y}_{i}^{(n+1)})-\phi(\hat{y}_{i}^{(n)})||_2^2.
\end{equation}

The total loss for fine-tuning enhancer model is the weighted combination of losses mentioned above:
\begin{equation}
\ell_{\mathrm{total}} = \ell_{\mathrm{con}} + \lambda_{\mathrm{R}}\ell_{\mathrm{R}},
\end{equation}
where $\lambda_{R}$ is the coefficient to balance different losses.
\section{Experiments}
\label{sec:Experiments}

\subsection{Experimental Settings}
\noindent \textbf{Training Details.} We implement our method with PyTorch framework on a single NVIDIA RTX 4090 GPU. During the training process of ranker model, we set the batch size as 8 and use the Adam optimizer~\cite{adam} with weight decay of $1\times10^{-4}$. In the model initialization phase, we first train the NIQE-oriented ranker model with an initial learning rate of $1\times10^{-4}$, which is halved every 20$K$ iterations for a total of 60$K$ iterations. The margin parameter $\epsilon$ in the margin-ranking loss is set to $0.5$. Subsequently, for each iterative training cycles, all human-oriented ranker models are trained with a fixed learning rate of $1\times10^{-5}$ for 5$K$ iterations. In the dense data selection process, the number of selected image pairs $k$ is set to 300.
For the fine-tuning process of the enhancer model, we employ a batch size of two and apply the Adam optimizer~\cite{adam} with a learning rate of $1\times10^{-5}$ for 10$K$ iterations. The loss balancing parameter $\lambda_{R}$ is set to 0.1.
We perform a total of $N=5$ iterative training stages in our HiLLIE framework.
All ranking data and collected labels which encode the annotators' preferences and models will be released to facilitate reproducibility.

\noindent \textbf{Datasets and Metrics.} We evaluate the proposed HiLLIE training framework on two commonly used paired LLIE datasets: LOLv1~\cite{super_re1} and LOLv2-real~\cite{lolv2}. 
Specifically, to facilitate unpaired training, we utilize all 485 low-light input images from LOLv1~\cite{super_re1} dataset along with randomly selected 485 normal-light ground-truth images from LSRW~\cite{LSRW} dataset for training, followed by evaluation on the 15-image paired test set of LOLv1~\cite{super_re1} dataset. 
For LOLv2-real~\cite{lolv2} dataset, we divide the original train set into two equal portions, utilizing 344 low-light input images and non-paired 344 normal-light ground-truth images for training, then evaluate on the 100-image paired test set.
Additionally, we conduct evaluations on three real-world unpaired benchmarks that only contain low-light images, including LIME~\cite{lime}, NPE~\cite{NPE}, and DICM~\cite{DICM}.

It is generally believed that no-reference IQA metrics focus more on overall image quality rather than pixel-wise differences from reference images, making them more aligned with human visual perception compared to full-reference IQA metrics. 
Based on this consideration, we adopt no-reference IQA metrics (PI~\cite{PI}, NIQE~\cite{NIQE}, and MUSIQ~\cite{MUSIQ}) as our major evaluation metrics, and provide full-reference IQA metrics (PSNR and SSIM~\cite{SSIM}) for reference. In prior research, PI was adopted in~\cite{zeroref_zero-DCE++,unsuper_lightendiffusion}, NIQE was utilized in~\cite{unsuper_enlightengan,zeroref_sci,unsuper_lightendiffusion}, and MUSIQ was employed in~\cite{unsuper_clip_lit}.
For LOLv1~\cite{super_re1} and LOLv2-real~\cite{lolv2} datasets with paired test images, we employ both types of IQA metrics for comprehensive evaluation. For the unpaired real-world sets, we only utilize no-reference IQA metrics due to the absence of ground-truth images.
All quantitative metrics are computed using the IQA-PyTorch Toolbox~\cite{pyiqa}.

\begin{table*}[htp]
\begin{center}
\caption{Quantitative comparisons on LOLv1~\cite{super_re1}, LOLv2-real~\cite{lolv2}, and unpaired LIME~\cite{lime}, NPE~\cite{NPE}, and DICM~\cite{DICM} datasets. `RetNet', `URetNet', `RetFmr', `EnGAN' and `LiDiff' are abbreviations for Retinex-Net~\cite{retinexnet}, URetinex-Net~\cite{super_re4_uretinexnet}, Retinexformer~\cite{super_retinexformer}, EnlightenGAN~\cite{unsuper_enlightengan} and LightenDiffusion~\cite{unsuper_lightendiffusion} methods, respectively. The best result of each category is highlighted in \textbf{bold}.
\vspace{-1.5mm}
}
\scalebox{0.69}{  
\setlength{\tabcolsep}{1.5pt}
\renewcommand{\arraystretch}{1}
\newcolumntype{C}[1]{>{\centering\arraybackslash}p{#1}}
\begin{tabular}{@{}C{1.35cm}C{1.5cm}C{1.7cm}C{1.7cm}C{1.7cm}C{1.7cm}C{1.9cm}C{1.7cm}C{1.7cm}C{1.7cm}C{1.75cm}C{1.7cm}C{1.7cm}C{1.7cm}@{}}
\toprule
\multirow{2}{*}{\textbf{Datasets}} & \multirow{2}{*}{\textbf{Metrics}} & \multicolumn{3}{c}{\textbf{Supervised Methods}} & \multicolumn{9}{c}{\textbf{Unsupervised Methods}} \\
\cmidrule(r){3-5} \cmidrule(l){6-14}
& & RetNet & URetNet & RetFmr & EnGAN & Zero-DCE++ & RUAS & SCI & PairLIE & CLIP-LIT & QuadPrior & LiDiff & Ours \\ 
\midrule
\multirow{5}{*}{\centering LOLv1}
& PSNR$\uparrow$ & 16.774 & 21.328 & \textbf{25.153} & 16.461 & 14.194 & 16.405 & 14.784 & 19.514 & 16.224 & 18.773 & \textbf{20.245} & 17.837 \\
& SSIM$\uparrow$ & 0.536 & 0.885 & \textbf{0.898} & 0.707 & 0.619 & 0.704 & 0.617 & 0.747 & 0.605 & 0.833 & \textbf{0.872} & 0.856 \\
& PI$\downarrow$ & 4.955 & 3.372 & \textbf{3.097} & 3.666 & 4.451 & 4.565 & 4.562 & 5.772 & 4.641 & 3.905 & 3.582 & \textbf{3.403} \\
& NIQE$\downarrow$ & 8.872 & 4.259 & \textbf{3.471} & 4.299 & 7.823 & 6.342 & 7.873 & 6.685 & 8.148 & 4.901 & 4.004 & \textbf{3.891} \\
& MUSIQ$\uparrow$ & 57.257 & \textbf{72.892} & 63.146 & 57.967 & 54.305 & 59.344 & 57.118 & 56.111 & 57.318 & 61.956 & 63.476 & \textbf{64.229} \\
\midrule
\multirow{5}{*}{\centering \begin{tabular}[c]{@{}c@{}}LOLv2-\\[-1pt]real\end{tabular}}
& PSNR$\uparrow$ & 16.097 & 21.222 & \textbf{22.794} & 14.929 & 12.581 & 15.326 & 17.304 & 18.256 & 16.673 & 20.444 & \textbf{22.682} & 22.223 \\
& SSIM$\uparrow$ & 0.512 & \textbf{0.878} & 0.867  & 0.656 & 0.539 & 0.673 & 0.629 & 0.774 & 0.604 & 0.838 & \textbf{0.876} & 0.874 \\
& PI$\downarrow$ & 5.269 & 3.193 & \textbf{3.031} & 3.064 & 4.686 & 4.401 & 4.602 & 5.424 & 4.752 & 3.602 & 3.138 & \textbf{2.973} \\
& NIQE$\downarrow$ & 9.426 & 4.384 & \textbf{3.967} & 4.787 & 8.155 & 6.533 & 8.047 & 6.794 & 8.414 & 4.910 & 3.820 & \textbf{3.707} \\
& MUSIQ$\uparrow$ & 56.672 & \textbf{69.495} & 59.304 & 56.030 & 52.009 & 56.177 & 54.443 & 53.460 & 54.830 & 58.201 & 59.571 & \textbf{60.673} \\
\midrule
\multirow{3}{*}{\centering LIME}
& PI$\downarrow$ & \textbf{3.080} & 3.219 & 3.124 & 3.184 & 3.344 & 3.923 & 3.172 & 3.269 & 3.267 & 4.441 & 3.231 & \textbf{3.139} \\
& NIQE$\downarrow$ & 4.598 & 4.392 & \textbf{4.069} & 4.255 & 4.687 & 5.372 & 4.180 & 4.240 & 4.417 & 5.565 & 4.058 & \textbf{3.950} \\
& MUSIQ$\uparrow$ & 62.905 & \textbf{65.003} & 60.840 & 58.910 & 52.610 & 55.424 & 59.113 & 59.199 & 57.004 & 56.720 & 58.849 & \textbf{59.208} \\
\midrule
\multirow{3}{*}{\centering NPE}
& PI$\downarrow$ & 3.161 & 2.961 & \textbf{2.845} & 3.244 & 3.630 & 4.901 & 3.204 & 3.006 & 4.492 & 3.413 & 2.717 & \textbf{2.688} \\
& NIQE$\downarrow$ & 4.520 & 3.869 & \textbf{3.489} & 3.976 & 4.695 & 6.739 & 4.629 & 3.889 & 6.435 & 4.210 & 3.232 & \textbf{3.179} \\
& MUSIQ$\uparrow$ & 66.589 & \textbf{67.063} & 62.050 & 59.354 & 52.315 & 51.514 & 59.390 & 62.777 & 57.957 & 58.150 & 62.637 & \textbf{62.996} \\
\midrule
\multirow{3}{*}{\centering DICM}
& PI$\downarrow$ & 2.942 & 2.513 & \textbf{2.479} & 2.927 & 2.971 & 3.734 & 2.810 & 2.984 & 3.831 & 3.243 & 2.638 & \textbf{2.561} \\
& NIQE$\downarrow$ & 4.502 & 3.325 & \textbf{3.098} & 3.534 & 3.427 & 4.717 & 3.132 & 3.563 & 5.119 & 4.220 & 3.172 & \textbf{3.031} \\
& MUSIQ$\uparrow$ & 64.506 & \textbf{65.289} & 59.344 & 58.418 & 50.912 & 53.828 & 58.283 & 58.627 & 56.712 & 59.126 & 60.396 & \textbf{60.512} \\
\bottomrule
\end{tabular}
}
\label{tab:sota_results}
\end{center}
\end{table*}

\begin{figure*}[t]
    \centering
    \includegraphics[width=0.99\linewidth]{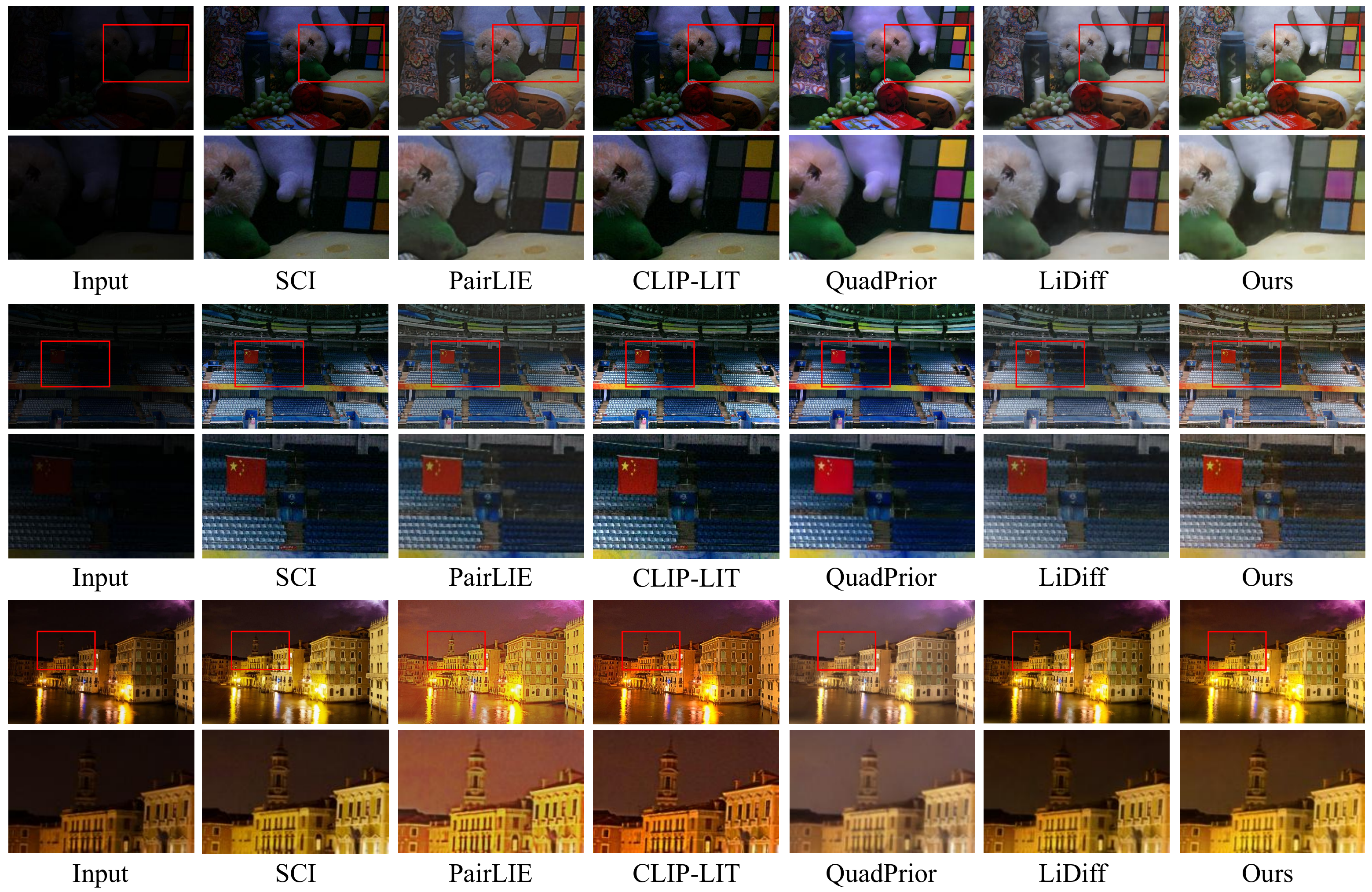}
    \caption{
Qualitative comparisons on LOLv1 (top row), LOLv2-real (middle row) and LIME datasets (bottom row).
}
    \label{fig:sota}
\end{figure*}

\subsection{Comparison with Existing Methods}
\label{sec:compare_sota}
We compare our model with two categories of LLIE methods: 1) three supervised methods including Retinex-Net~\cite{retinexnet}, URetinex-Net~\cite{super_re4_uretinexnet}, and Retinexformer~\cite{super_retinexformer}; 2)
eight unsupervised methods, among which EnlightenGAN~\cite{unsuper_enlightengan}, PairLIE~\cite{unsuper_PairLIE}, CLIP-LIT~\cite{unsuper_clip_lit},  and LightenDiffusion~\cite{unsuper_lightendiffusion} are unsupervised methods trained using unpaired low- and normal-light images, while Zero-DCE++~\cite{zeroref_zero-DCE++}, RUAS~\cite{zeroref_RUAS}, SCI~\cite{zeroref_sci}, and QuadPrior~\cite{zeroref_quadprior} are zero-reference methods which only utilize low- or normal-light images for training.
Note that the supervised approaches are evaluated without retraining, as our method does not require paired training data.
All unsupervised methods are retrained using the same data as our method for fair comparison.

\noindent \textbf{Quantitative Comparison.} 
The quantitative results of different LLIE methods are presented in Tab.~\ref{tab:sota_results}.
Although the baseline model, LightenDiffusion~\cite{unsuper_lightendiffusion}, achieves high pixel-level fidelity scores, it lacks advantages in no-reference IQA metrics that better reflect perceptual quality across all datasets.
With our proposed HiLLIE framework, we observe noticeable improvements in no-reference IQA metrics, outperforming other unsupervised LLIE methods and even some supervised approaches.
Nevertheless, since our method prioritizes outputs better aligned with human perception, this inevitably leads to a trade-off in pixel-level fidelity, resulting in lower PSNR and SSIM scores compared to the baseline model. 
The results on unpaired real-world datasets continue to show the effectiveness of our method, demonstrating its generalization capability in real-world scenarios.

\noindent \textbf{Qualitative Comparison.} 
Visual results of different unsupervised LLIE methods are shown in Fig.~\ref{fig:sota}. 
Compared to the baseline method and other competitive LLIE approaches, our proposed method achieves superior performance in enhancing global illumination while preserving local contrast, without introducing over- or under-exposure artifacts across different lighting conditions.
This balanced enhancement improves overall image visibility and successfully reveals previously obscured details in dark regions, as evidenced by the identifiable seats in the upper right area of the middle row image and the architectural details of the tower in the upper left region of the bottom row image.
Furthermore, our results exhibit more vivid and naturally appearing colors without introducing noticeable color distortions, such as the toys and color charts displayed in the top row image.
More visual results could be found in the supplementary material.

\subsection{Ablation Studies}
\label{sec:ablation}
\begin{figure*}[t]
\scalebox{0.99}{ 
\begin{minipage}{0.68\linewidth}
    \centering
    \includegraphics[width=1\linewidth]{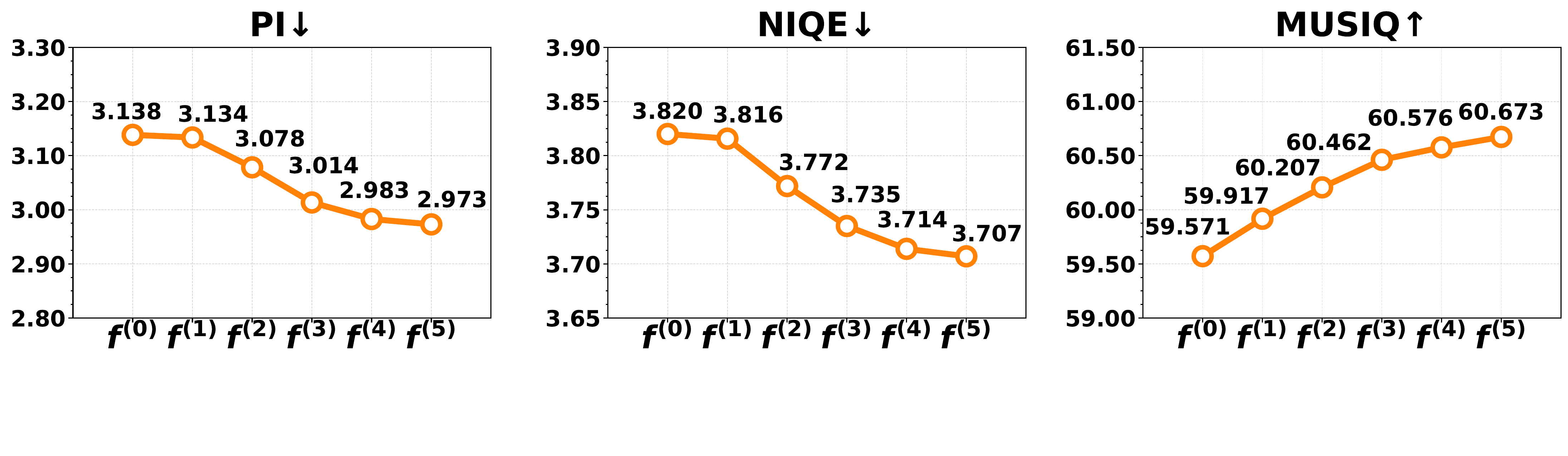}
    \vspace{-6mm}
    \caption{Ablation study of enhancer model performance on no-reference IQA metrics across different training stages. 
    More details can be found in Sec.~\ref{sec:ablation}.
    }
    \label{fig:cycle_ablation}
\end{minipage}%
\quad
\begin{minipage}{0.3\linewidth}
    \centering
    \vspace{4.5mm}
    \includegraphics[width=1\linewidth]{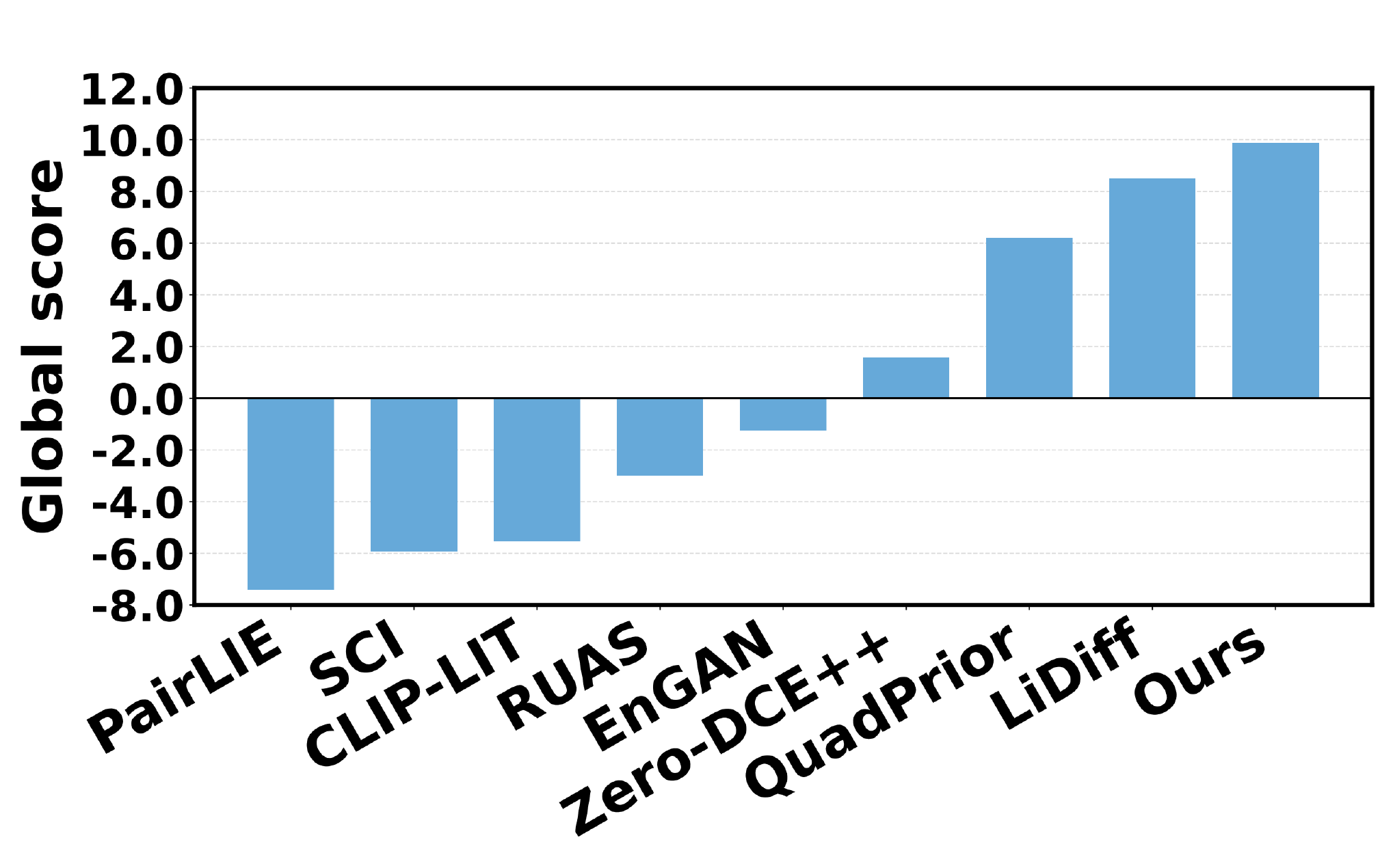}
    \vspace{-5.5mm}
    \caption{User Study. Global scores of different LLIE methods. 
   More details can be found in Sec.~\ref{sec:user study}.
   }
    \label{fig:MOS}
\end{minipage}}
\end{figure*}

\noindent \textbf{Human-in-the-Loop Iterative Training Cycles.} 
Within our proposed HiLLIE training framework, the enhancer model improvement is achieved through alternating training of ranker and enhancer models rather than a single-stage fine-tuning process. 
Our analysis focuses on no-reference IQA metrics, which better correlate with human perception of image quality compared to full-reference IQA metrics. 
As shown in Fig.~\ref{fig:cycle_ablation}, we observe consistent improving trends of no-reference IQA metrics across different training stages, demonstrating progressive improvements of enhancer performance throughout the iterative process.
By requiring only a small amount of human annotation effort at each stage, we progressively improve ranker's capability to simulate human visual assessment of enhancer outputs. This enables ranker to guide the enhancer optimization through a ranker loss, leading to results with more natural illumination levels and vivid colors that better align with human visual preferences for high-quality normal-light images.

\noindent \textbf{Effect of Ranker Loss.}  
To comprehensively evaluate the effectiveness of ranker loss $\ell_{R}$ and investigate the impact of different ranker training strategies, 
\begin{table}[t]
\begin{center}
\caption{Effect of ranker loss on enhancer model performance. 
Specifically, ``w/o $\ell_{R}$" indicates the model trained only with $\ell_{{con}}$, ``w $\ell_{R}, g^{(0)}$" denotes the model trained with $\ell_{{con}}$ and $\ell_{R}$ calculated by $g^{(0)}$, and ``w $\ell_{R}, g^{(n)}$" represents the model trained with $\ell_{{con}}$ and $\ell_{R}$ calculated by $g^{(n)}$ where $n=1,\ldots,5$. The best results are highlighted in \textbf{bold}. More details can be found in Sec.~\ref{sec:ablation}.
}
\vspace{-1.5mm}
\scalebox{0.75}{
\setlength{\tabcolsep}{1.0pt} 
\renewcommand{\arraystretch}{1}  
\newcolumntype{C}[1]{>{\centering\arraybackslash}p{#1}}
\begin{tabular}{@{}C{1.5cm}C{1.8cm}C{2.2cm}C{2.2cm}C{2.2cm}@{}}  
\toprule
\multirow{2}{*}{\textbf{Dataset}} & \multirow{2}{*}{\textbf{Metrics}} & \multicolumn{3}{c}{\textbf{Training Settings}} \\
\cmidrule(r){3-5}
& & w/o $\ell_{R}$ & w $\ell_{R}, g^{(0)}$ & w $\ell_{R}, g^{(n)}$ \\ 
\midrule
\multirow{5}{*}{\centering \begin{tabular}[c]{@{}c@{}}LOLv2-\\[-1pt]real\end{tabular}}
& PSNR$\uparrow$ & \textbf{22.856} & 22.745 & 22.067 \\
& SSIM$\uparrow$ & \textbf{0.878} & 0.877 & 0.874 \\
& PI$\downarrow$ & 3.159 & 3.093 & \textbf{2.930} \\
& NIQE$\downarrow$ & 3.822 & 3.792 & \textbf{3.697} \\
& MUSIQ$\uparrow$ & 59.365 & 60.030 & \textbf{61.008} \\ 
\bottomrule
\end{tabular}
}
\vspace{-5mm}
\label{tab:ranker_ablation}
\end{center}
\end{table}
we conduct experiments with three variants: (1) a baseline enhancer model trained without $\ell_{R}$, (2) an enhancer model guided by $\ell_{R}$ calculated using NIQE-oriented ranker model $g^{(0)}$, and (3) our full enhancer model guided by $\ell_{R}$ calculated using human-oriented ranker model $g^{(n)}$.
All models are trained for the same number of iterations for fair comparison.
Through these experiments, we aim to validate both the necessity of $\ell_{R}$ and the superiority of our iteratively trained human-oriented ranker model over the NIQE-oriented one in simulating human visual assessment of enhancer outputs.
As shown in Tab.~\ref{tab:ranker_ablation}, both models fine-tuned with $\ell_{R}$ calculated by $g^{(0)}$ or $g^{(n)}$ outperform the baseline model trained without $\ell_{R}$ across all no-reference metrics, indicating that the introduction of $\ell_{R}$, rather than increased model training time, plays an important role in improving the visual quality of enhanced images.
Moreover, the model fine-tuned with $\ell_{R}$ calculated by $g^{(n)}$ has better performance than the model fine-tuned using $g^{(0)}$.
This validates that our iteratively trained human-oriented ranker model, benefiting from incrementally collected quality labels encoded with human visual preference information, can better simulate human visual assessment of enhancer outputs and provide effective perceptual guidance for the enhancer training process.

\subsection{User Study}
\label{sec:user study}
Through iterative training of ranker and enhancer models, our ultimate goal is to generate enhanced results with better visual quality and perceptually preferred characteristics.
Following the subjective evaluation settings adopted in IQA literature~\cite{user1,user2}, we perform a user study to assess human preferences for enhanced outputs from all compared unsupervised LLIE approaches, which provides more direct human perception feedback on image perceptual quality compared to quantitative metrics.
Considering the potential evaluator fatigue, we refer to the settings of \cite{unsuper_clip_lit, user_study_refer} and select the first 40 consecutive low-light images in numerical order of original indices from LOLv2-real~\cite{lolv2} test set, which were then processed using each compared model to obtain enhanced images for comparison.
We invite 15 participants to evaluate the visual quality of generated images following the two-alternative forced choice (2AFC) method.
In particular, all the samples are first anatomized and randomized in order to ensure fairness of the survey.
During the evaluation process, participants are presented with a side-by-side comparison of enhanced results generated by two different methods using the same input image, and are asked to select the one with higher quality.
Once all subjective evaluations are completed, we organize the collected human preference data into a 9× 9 matrix $C$, where each entry $C_{i,j}$ represents the number of times enhanced results from method $i$ are preferred over those from method~$j$.
Finally, we employ the Thurstone's model~\cite{thurstone} with maximum likelihood estimation to compute the global scores.
This is achieved by finding the optimal global score solution $q=[q_1,q_2,...,q_9]$ for maximizing the log-likelihood of the quality preference matrix $C$.

The final global scores for compared unsupervised LLIE methods are presented in Fig.~\ref{fig:MOS}.
Larger global score indicates better image perceptual quality as perceived by human evaluators.
The statistics show that our results are the most favored by participants compared to other methods.
This serves as a strong support to demonstrate that our HiLLIE training framework could generate human-desired enhanced results by incorporating human visual preferences into the model training process with only small-scale annotation cost.
We also provide another user study conducted only by label annotators in the supplementary material, which further verifies that our model outputs meet the aesthetic preferences of annotators involved in the HiLLIE training framework.

\section{Conclusion}
\label{sec:conclusion}

We have presented a new human-in-the-loop low-light image enhancement training framework for unsupervised LLIE models.
Our approach successfully integrates human preferences through an iterative training between ranker and enhancer models.
With only small-scale annotation effort required at each stage, we update ranker to better simulate human visual assessment of enhancer outputs, and then integrate the learned human perceptual preferences into enhancer training process through ranker loss.
Extensive experiments validate the effectiveness of our approach in improving unsupervised LLIE model performance and producing enhanced results that better align with human visual perception.
Notably, our approach is the first attempt to apply the idea of human-in-the-loop into the training process of unsupervised LLIE models, and we will continue to explore its effectiveness on other tasks such as image tone mapping.

{
    \small
    \bibliographystyle{ieeenat_fullname}
    \bibliography{main}
}

\appendix
\clearpage
\setcounter{page}{1}
\setcounter{section}{0} 
\renewcommand{\thesection}{\Alph{section}}
\renewcommand{\thesubsection}{\Alph{section}.\arabic{subsection}}
\maketitlesupplementary
In this supplementary material, we provide more details and experiments of our proposed HiLLIE framework. 
First, we show examples of ranking dataset with corresponding labels across different training stages (Sec.~\ref{sec:labels}). 
We then discuss our mitigation strategy towards potential preference bias during human annotation process and validate the effectiveness of our approach through expanded user evaluation (Sec.~\ref{sec:bias}). 
Next, we verify the alignment of our enhancement direction with label annotators' preferences through additional experimental validations (Sec.~\ref{sec:verification}). 
We also illustrate our user study interface (Sec.~\ref{sec:user study interface}) and present the network architecture of our image quality ranker (Sec.~\ref{sec:network}). 
Finally, we provide more visual comparisons of state-of-the-art unsupervised LLIE methods and our methods (Sec.~\ref{sec:more visual results}).

\section{Examples of Ranking Dataset Collection}
\label{sec:labels}
In our experiments, we conduct HiLLIE training for $N=5$ iterative stages and obtain four human-oriented ranker models ($g^{(1)}$ to $g^{(4)}$).
During each stage $n$, we employ a dense data selection approach that identifies top-$k$ image pairs exhibiting the largest discrepancy in quality scores predicted by the ranker model $g^{(n-1)}$. 
This strategy enables efficient human annotation by focusing on the most representative samples in each stage, significantly reducing the annotation workload while maintaining the effectiveness of the training process.
Fig.~\ref{fig:label} presents the top-10 image pairs in set $\mathcal{S}^{(n)}$ along with their corresponding $g^{(n-1)}$ predicted value differences $\delta$ within each pair. For illustration purposes, we only show top-10 pairs here, while in practice we select top-300 pairs in each stage during training.
The $\delta$ values within each pair reveals a consistent trend across four stages where both the maximum and minimum values of $\delta$ gradually decrease as $n$ increases.
This trend indicates that the difference in visual quality between enhancer outputs from adjacent stages progressively reduces, suggesting both the improvement and convergence of the enhanced results in terms of human visual preference.
During the annotation process, we instruct human annotators to select images based on their personal visual preferences rather than similarity to ground-truth images, as no ground-truth images are provided during training.
To visualize the rank labeling results, we use checkmarks to indicate the images that human annotators deem of higher visual quality, and crosses for counterparts.

\section{Preference Bias Discussion}
\label{sec:bias}
One of the major concern of human annotation may be potential preference bias across different individuals, which indeed is a commonly observed phenomenon.
We have considered this preference bias issue during the data annotation process. 
To increase the reliability of annotated labels, as mentioned in Sec.~3.2.1, we employ 3 annotators and adopt majority voting for the final ranking labels.
The choice of number of annotators is the same as that of other tasks such as~\cite{human_T2I, troubleshooting}.
In order to validate our model performance with a broader audience, as mentioned in Sec.~4.4, we invite 15 participants to evaluate the visual quality of enhancement results from different unsupervised LLIE methods.
According to Fig.~4 in Sec.~4.4, these 15 evaluators generally prefer our model outputs.
Since the number of user study evaluators exceeds that of annotators, this suggests that our method has learned some general preferences and shows generalization ability across different individuals.
Notably, as mentioned in our contributions, our method primarily provides an algorithmic implementation that directly incorporates human visual preferences into the training process of an unsupervised LLIE model through a tailored IQA model.
Despite our efforts above, there might still exist potential bias due to the limited number of participants, which we will make further attempts to improve in future work.

\section{Verification of Enhancement Direction Alignment with Annotators' Preference}
\label{sec:verification}
In this section, we demonstrate from three aspects how our HiLLIE framework can progressively improve ranker model's capability to capture human visual preferences, and thereby guide the enhancer to generate LLIE outputs with human-desired characteristics that closely align with the visual preferences of label annotators.

\begin{figure}[t]
    \centering
    \includegraphics[width=0.99\linewidth]{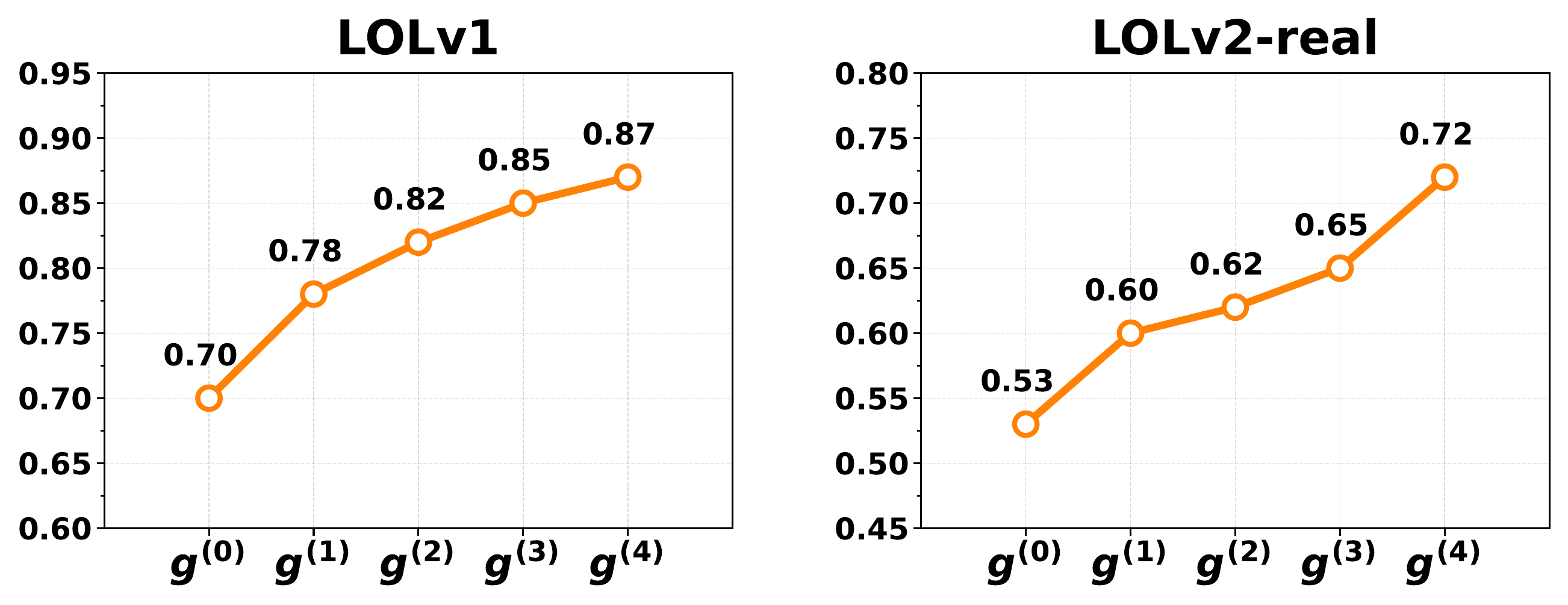}
    \vspace{-1.5mm}
    \caption{
    Prediction accuracy of $g^{(n)}$ on all collected $f^{(n)}$ outputs.
    }
    \label{fig:ranker better}
\end{figure}

First, to evaluate whether the ranker's ability to simulate human visual assessment of enhancer outputs increases with advancing stages $n$, using all enhancer ${\{f^{(n)}\}}_{n=0}^5$ outputs on the validation set as ranker ${\{f^{(n)}\}}_{n=0}^4$ inputs, we calculate the prediction accuracy between the predicted scores of $g^{(n)}$ and human annotated labels.
Specifically, given two images with identical content but produced by different $f^{(n)}$, $g^{(n)}$ predicts which one has superior visual quality. 
This prediction is then compared to human annotation result that indicates which image annotators prefer. 
The prediction accuracy is defined as the percentage of cases where the judgement of $g^{(n)}$ matches that of annotators regarding the relative quality relationship within image pairs. 
Fig.~\ref{fig:ranker better} shows the accuracy results of all $g^{(n)}$ calculated on the LOLv1 and LOLv2-real datasets.
It can be seen that as $n$ increases, the accuracy of $g^{(n)}$ grows accordingly, indicating improved ranker's ability to capture annotators' aesthetic preferences.

\begin{figure}[t]
    \centering
    \includegraphics[width=0.99\linewidth]{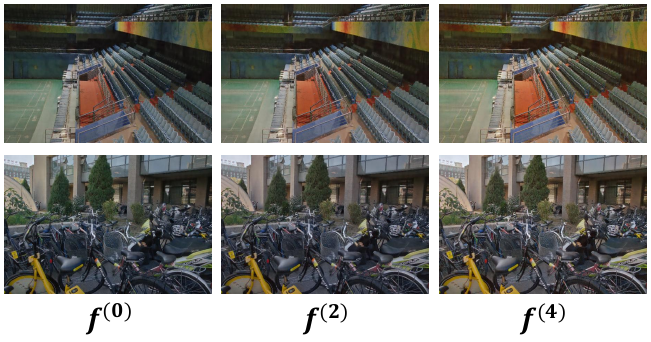}
    \vspace{-1.5mm}
    \caption{
Qualitative comparison between enhanced results of $f^{(n)}$ across different stages.
}
    \label{fig:diff stage}
\end{figure}

Next, we examine whether the enhancer outputs iteratively progress toward the direction that the ranker deems as having better visual quality.
As mentioned in Sec.~3.3, we employ ranker as a criterion to guide the training process of enhancer.
Since ranker can simulate human visual assessment of enhancer outputs with relatively high prediction accuracy, confirming that successive enhancer outputs consistently move in the direction favored by the ranker would indicate that our enhancement results are becoming progressively more aligned with annotators' visual preferences.
Specifically, for each $g^{(n)}$, we test its prediction accuracy on the enhanced outputs of $f^{(n)}$ and $f^{(n+1)}$ on the validation set.
Different from the previous accuracy calculation experimental setting, here we use pseudo labels instead of human annotated ones, assuming that $f^{(n+1)}$ outputs always have better visual quality than $f^{(n)}$ outputs. 
In this case, an accuracy greater than 0.5 would demonstrate that $g^{(n)}$ considers the average visual quality of $f^{(n+1)}$ outputs to be superior to that of $f^{(n)}$ outputs.
Tab.~\ref{tab:ranker_prediction} shows the prediction accuracy results under this experimental setup.
It can be observed that all $g^{(n)}$ evaluate the outputs of $f^{(n+1)}$ as having better visual quality than those of $f^{(n)}$, confirming that the enhancer evolves in a direction aligned with ranker preferences, and thus produces perceptually improved results that match the visual preferences of label annotators.
%
As shown in Fig.~\ref{fig:diff stage}, we visualize the enhanced outputs of $f^{(n)}$ across different stages $n$. As $n$ increases, we observe continuous improvements in the visual quality of enhanced results. 
The images progressively exhibit perceptually superior qualities, featuring brighter illumination and more natural colors, which better align with human visual preferences.
\begin{table}[t]
\begin{center}
\caption{Prediction accuracy of $g^{(n)}$ on $f^{(n)}$ and $f^{(n+1)}$.}
\vspace{-1.5mm}
\scalebox{0.8}{
\setlength{\tabcolsep}{0.5pt}  
\renewcommand{\arraystretch}{1}     
\newcolumntype{C}[1]{>{\centering\arraybackslash}p{#1}}
\begin{tabular}{@{}C{4cm}C{1.8cm}C{2.7cm}@{}}  
\toprule
\multirow{2}{*}{\textbf{Models}} & \multicolumn{2}{c@{}}{\textbf{Datasets}} \\
\cmidrule(r){2-3}
& LOLv1 & LOLv2-real \\ 
\midrule
$g^{(0)}$ on $f^{(0)}$ and $f^{(1)}$ & 0.77 & 0.77 \\
$g^{(1)}$ on $f^{(1)}$ and $f^{(2)}$ & 0.77 & 0.76 \\
$g^{(2)}$ on $f^{(2)}$ and $f^{(3)}$ & 0.77 & 0.73 \\
$g^{(3)}$ on $f^{(3)}$ and $f^{(4)}$ & 0.87 & 0.75 \\
$g^{(4)}$ on $f^{(4)}$ and $f^{(5)}$ & 0.80 & 0.82 \\
\bottomrule
\end{tabular}
}
\vspace{-3mm}
\label{tab:ranker_prediction}
\end{center}
\end{table}

\begin{figure}[t]
    \centering
    \includegraphics[width=0.99\linewidth]{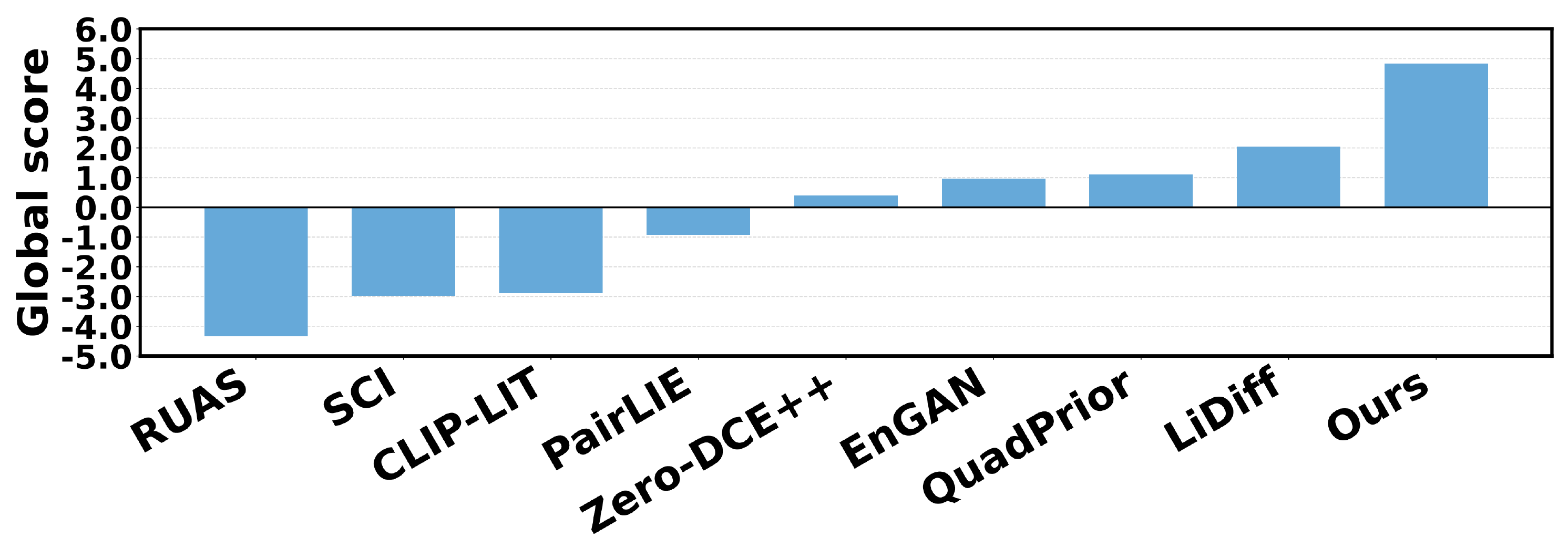}
    \vspace{-1.5mm}
    \caption{
    Global scores of user study conducted by annotators.
    }
    \label{fig:annotator user study}
\end{figure}

Furthermore, we conduct an additional user study with the same annotators to directly validate the alignment between our model outputs and visual preferences of label annotators. 
As shown in Fig.~\ref{fig:annotator user study}, when comparing our model outputs with those from other unsupervised LLIE methods, the annotators consistently favored our results by a significant margin.
This strong consensus among annotators serves as compelling evidence that our enhancement results successfully align with annotators' visual expectations, validating the effectiveness of our HiLLIE framework.

\section{User Study Interface}
\label{sec:user study interface}
\begin{figure*}[t]
    \centering
    \includegraphics[width=0.99\linewidth]{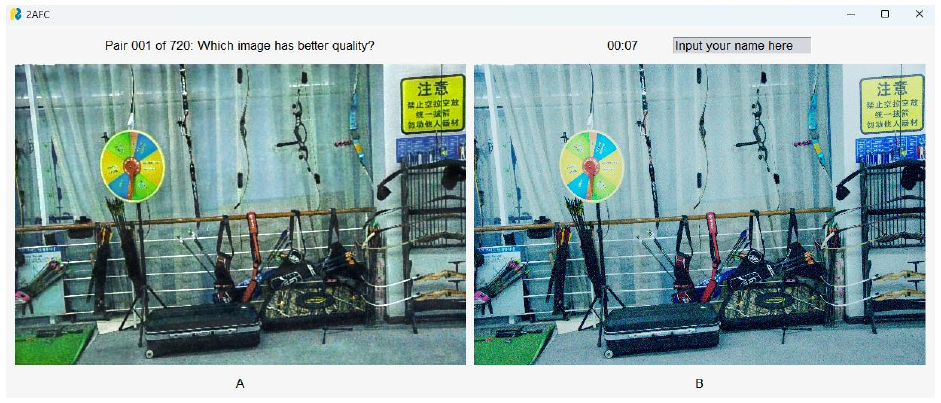}
    \caption{
Interface for conducting pairwise comparisons in our user study.
}
    \label{fig:interface}
\end{figure*}

Fig.~\ref{fig:interface} illustrates the interface utilized for conducting pairwise comparisons between enhanced results from different LLIE methods. During the evaluation, human subjects are presented with two images of the same content (A and B) and requested to click on the one with higher perceptual quality. To facilitate detailed assessment, subjects can use mouse scroll to zoom in and examine image details. After each selection, the system automatically records the choice and advances to the next pair, while the interface header tracks both current progress and time spent. The complete evaluation comprises 1440 image pairs, typically requiring 60-80 minutes per subject to complete. Upon completion, all selection results are automatically saved to a CSV file for subsequent statistical analysis.

\section{Image Quality Ranker Architecture}
\label{sec:network}
\begin{figure*}[t]
    \centering
    \includegraphics[width=0.99\linewidth]{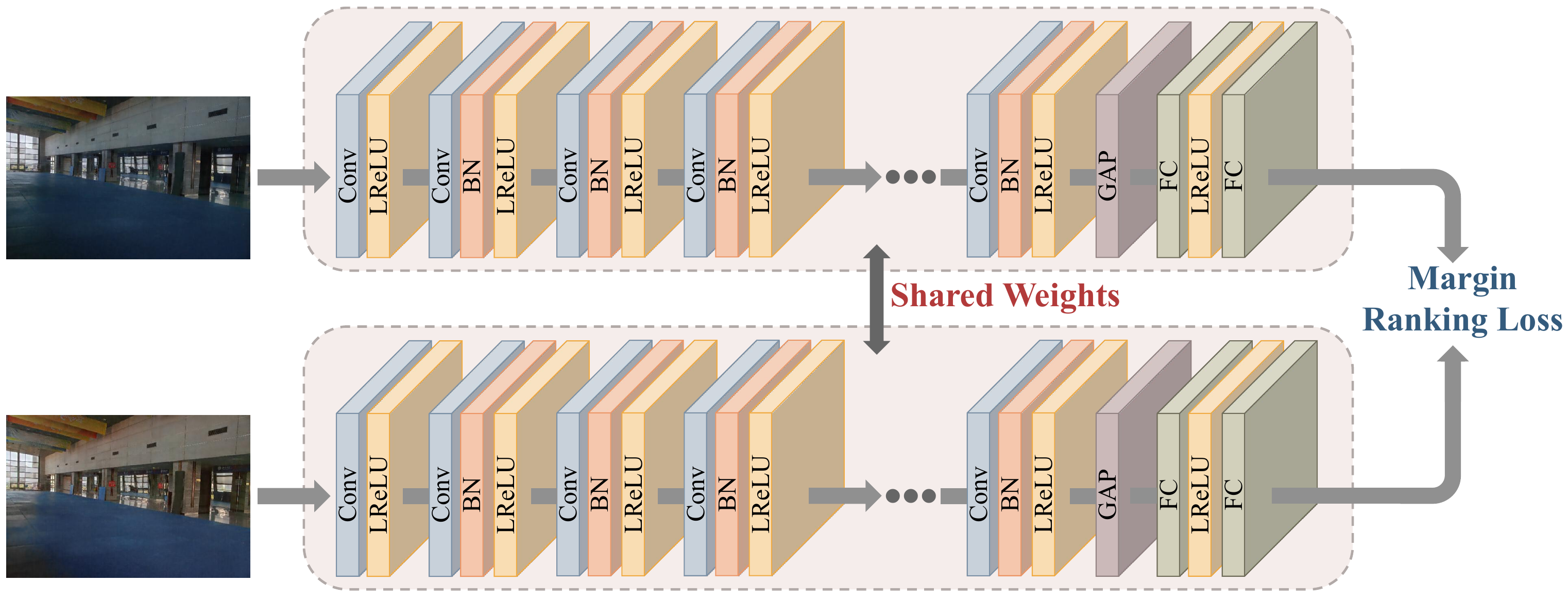}
    \caption{
The network architecture of our image quality ranker.
}
    \label{fig:ranker arch}
\end{figure*}

\begin{figure*}[t]
    \centering
    \includegraphics[width=0.99\linewidth]{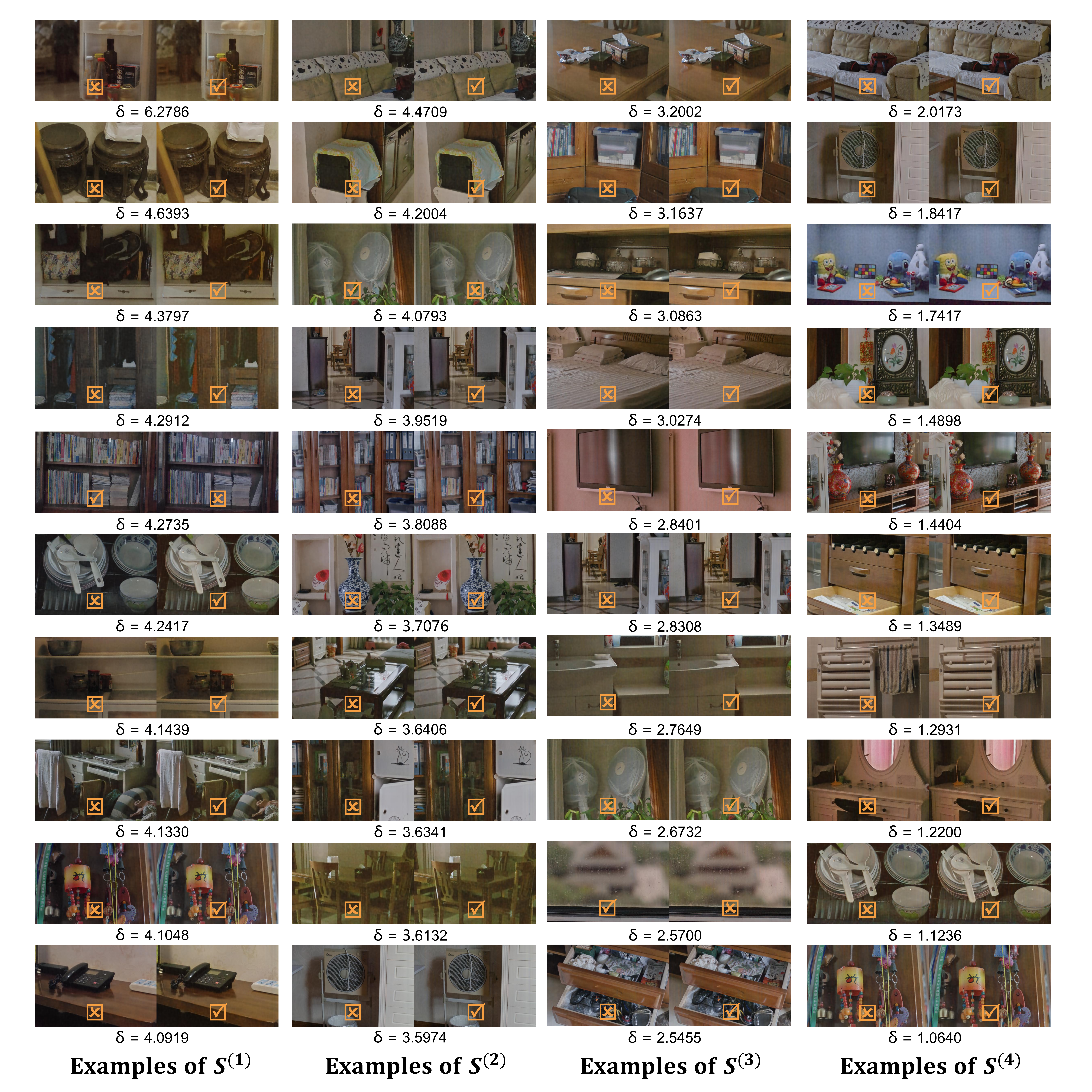}
    \caption{
Selected image pair examples in $\mathcal{S}^{(n)}$ with their corresponding rank labeling results.
}
    \label{fig:label}
\end{figure*}

Our ranker adopts a siamese-like architecture~\cite{siamese, ranksrgan}, where two identical network branches with shared weights are used to process the paired images independently.
As shown in Fig.~\ref{fig:ranker arch}, each branch starts with an initial convolutional layer with LeakyReLU activation, followed by a sequence of nine identical convolutional blocks.
Each block consists of a convolutional layer, batch normalization, and LeakyReLU activation in sequence.
After the convolutional blocks, a global average pooling layer aggregates the spatial information, which are then feed into two fully connected layers with LeakyReLU activation in between.
Finally, each branch outputs a scalar value representing the predicted quality score of the corresponding image, which is used to compute the margin ranking loss.


\section{More Visual Results}
\label{sec:more visual results}
\begin{figure*}[t]
    \centering
    \includegraphics[width=0.99\linewidth]{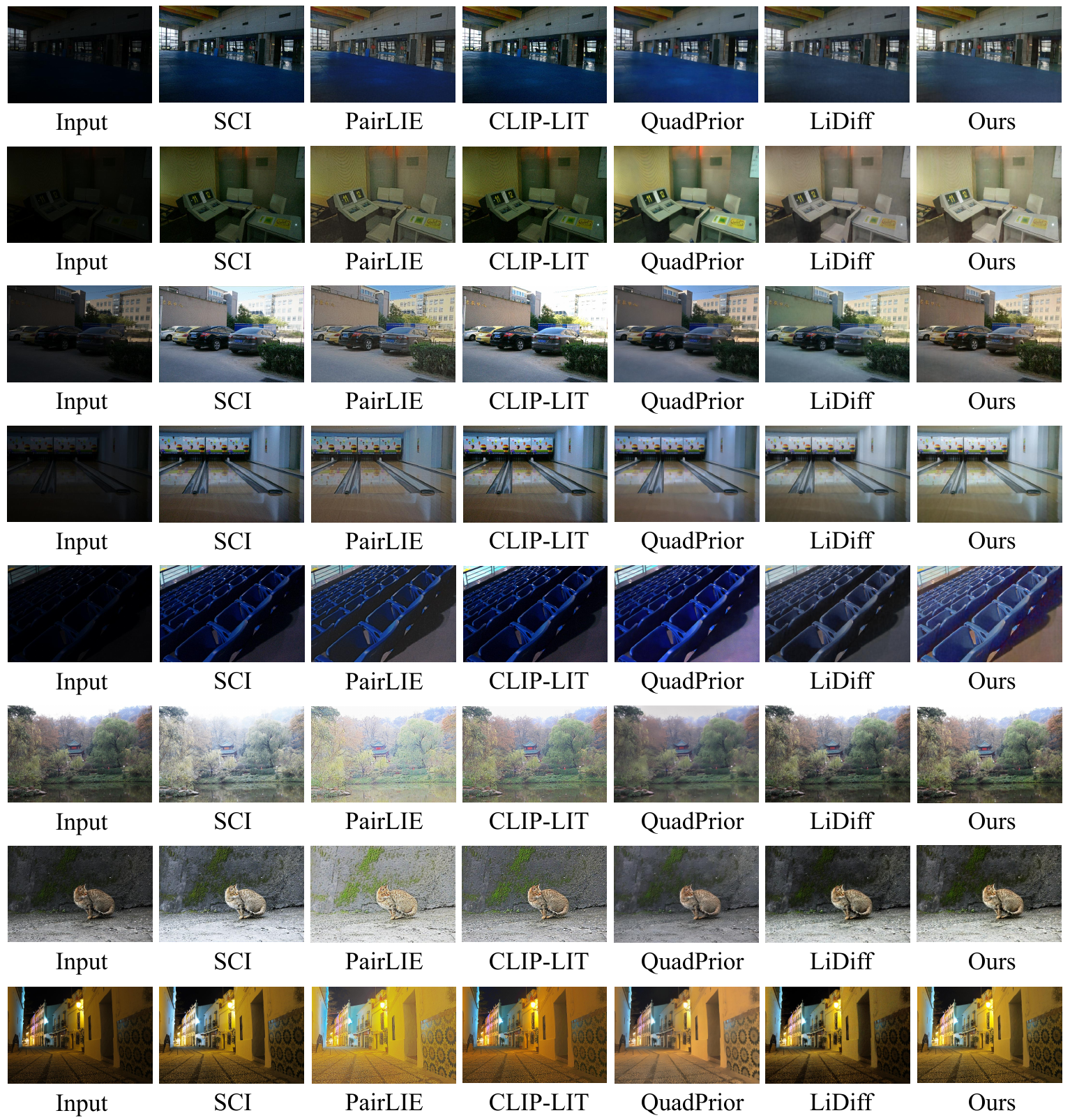}
    \caption{
More visual results of compared unsupervised LLIE methods and our method on benchmarks.
}
    \label{fig:more1}
\end{figure*}

As shown in Fig.~\ref{fig:more1}, we present more visual results of different LLIE methods including SCI~\cite{zeroref_sci}, PairLIE~\cite{unsuper_PairLIE}, CLIP-LIT~\cite{unsuper_clip_lit}, QuadPrior~\cite{zeroref_quadprior}, LightenDiffusion~\cite{unsuper_lightendiffusion} and our method. 
Compared with other methods, our approach generates superior enhanced results with improved illumination and vivid colors that better align with human perception.

\end{document}